\documentclass[runningheads]{llncs}
\usepackage[T1]{fontenc}
\usepackage{comment}
\usepackage{booktabs}
\usepackage{amsmath}
\usepackage[backref=page]{hyperref}
\usepackage{tabularx}
\newcolumntype{Y}{>{\centering\arraybackslash}X}
\usepackage{graphicx}
\usepackage{color}

\begin{document}
\title{Toward Foundation Detection Models via Example-Based Grounding in EM-DETR}
\titlerunning{Foundation Models with EM-DETR}
%
\author{Sheethal Bhat\inst{1,2}\orcidID{0009-0006-2044-5113} \and
Bogdan Georgescu\inst{3} \and
Awais Mansoor\inst{3} \and
Florin C. Ghesu\inst{2} \and
Sasa Grbic\inst{3} \and
Andreas Maier\inst{1}}
\authorrunning{S. Bhat et al.}

\institute{Friedrich-Alexander-Universität, 91058, Erlangen-Nürnberg, Germany \and
Siemens Healthineers, 91052, Erlangen, Germany
\and
Siemens Medical Solutions, 08540, Princeton, USA\\
\email{sheethal.bhat@fau.de}}
\maketitle              
\begin{abstract}
Deep Learning (DL)-based Computer-Aided Diagnostic (CAD) systems have shown great promise in supporting clinicians, but their performance heavily depends on the availability of large, well-annotated datasets. In medical lesion detection, the scarcity of comprehensive, standardized annotations and the heterogeneity of existing datasets limit the development of robust foundation models. In this work, we introduce a novel approach to building a multi-domain foundation model for lesion detection by integrating diverse medical image datasets annotated for both anatomical structures and abnormalities. We first propose a student-teacher training strategy to effectively combine datasets with heterogeneous label spaces, mitigating catastrophic forgetting, and improving feature learning. In addition, building on our previously proposed example-based EM-DETR framework, we adapt it to jointly learn from multiple domain-specific datasets, enabling fine-grained, example-driven detection across modalities. Our model achieves state-of-the-art (SOTA) results, with image-level AUC scores exceeding previous SOTA by 1\% point and on four key findings in CXRs.We achieve a leading $mAP_{50}$ in CXRs and mammography, surpassing EM-DETR trained on a single dataset by 2-3\% points. 

\end{abstract}

\section{Motivation}

Object detection is a fundamental challenge in computer vision, finding extensive use in fields such as autonomous driving, robotics, and surveillance. In medical imaging, detection plays an equally critical role, enabling the identification and localization of anatomical structures or pathological findings that guide clinical decision making. Traditional CNN-based detection models such as Faster R-CNN \cite{fasterrcnn}, RetinaNet \cite{li2019light}, and FCOS \cite{tian2020fcos} have demonstrated strong performance in natural image domains and have also been adapted for radiographic tasks \cite{ghosh2024mammo}. However, these models often rely on extensive, dataset-specific supervision and struggle to generalize across modalities or out-of-distribution (OOD) data \cite{emdetr}. This is primarily due to the intricate nature of obtaining high-quality annotations, the subtle visual indicators that complicate the defining of boundary regions, and the diversity of anatomical structures \cite{nguyen2022vindr,rueckel2021pneumothorax}.

To address the limitations of supervised object detectors, recent work has explored Vision-Language (VL) detection models \cite{li2022grounded,liu2024grounding,kim2023region,minderer2023scaling,minderer2022simple} that integrate image and text supervision during training. These models show strong potential for generalizing across novel classes and domains, particularly in natural image settings. Models such as GLIP \cite{li2022grounded}, OWL-ViT \cite{minderer2022simple}, and Grounding DINO \cite{liu2024grounding} achieve state-of-the-art detection performance by leveraging powerful, jointly learned visual and language embeddings, which are aligned through contrastive learning. Moreover, large-scale pretraining on “grounding” or “caption” datasets i.e., region-level textual descriptions, enables open-set zero-shot detection, allowing recognition of unseen classes from their textual cues. This pretraining leverages automatically mined image-caption pairs from extensive web sources and curated datasets, including LAION-400M \cite{schuhmann2021laion}, Visual Genome \cite{krishna2017visual}, and COCO Captions \cite{chen2015microsoft}. 

In contrast, the medical domain lacks comparable comprehensive image-text paired datasets, posing challenges for effective VL model training. Medical image annotations are sparse \cite{nguyen2022vindr,NguyenMammo}, domain-specific, and expensive to obtain \cite{ghesu2022contrastive}, often requiring expert knowledge and adherence to stringent annotation protocols. Consequently, the development of a similar foundational model for medical lesion detection continues to pose significant challenges. In this study, we initially examine two distinct datasets: annotated for anatomical structures and abnormalities, with the goal of enhancing the visual feature space during baseline training. While this approach offers modest gains, we extend it further by leveraging our previously proposed EM-DETR \cite{emdetr} framework. Specifically, we use EM-DETR \cite{emdetr}, an example-based multi-modal detector accepted at MICCAI 2025, to integrate multiple domain-specific datasets, allowing the model to learn fine-grained, example-based features tailored to distinct lesion types across various imaging contexts. 

Our contributions are twofold: (1) we explore combining anatomical and abnormality annotations to enrich supervision during training, and (2) we extend EM-DETR to develop a multi-domain, example-based foundation detection model that achieves state-of-the-art (SOTA) results across five key findings in CXRs and mammography. The remainder of the paper is organized as follows: Section 2 details the datasets used, covering both anatomical and abnormality annotations. Section 3 outlines the two methods of developing a foundation model. Section 4 presents experimental results, and Section 5 concludes with discussion and future directions.


\section{Datasets}
\label{sec:datasets}

This work utilises four image datasets across two modalities: Chest X-Rays (CXRs) and mammography. These include a private dataset annotated for key anatomical structures, two datasets with abnormality-level labels (private \cite{ghesu2022contrastive} and public \cite{nguyen2022vindr}), and an additional public mammography dataset \cite{NguyenMammo} annotated for abnormalities at the lesion level. The details of each dataset, including annotation types, and class distribution, are described below.  

\subsubsection*{Dataset $A, B$}
are two separate data subsets formed from a larger dataset acquired from multiple sites within the United States and Europe. This dataset includes frontal and lateral chest exams from adult cohorts ($>$18 years). For annotation, two general radiologists independently reviewed all CXRs for the presence and location of 16 findings. A third general radiologist served as an independent tiebreaker for discrepant findings. Dataset $\mathcal{A}$ consists of 15,000 full-resolution frontal view images with heart and lung annotations. Here we divide the lung annotations into right lung and left lung class. The dataset $\mathcal{B}$ consists of 16,953 full resolution frontal view images with bounding boxes that indicate the presence of 16 findings, provided by an expert panel of radiologists. We employ four expert radiologist-provided gold-standard datasets for evaluation. These test datasets are binary classified and annotated for the presence or absence of each finding. These include 333 images for both atelectasis and consolidation, 331 images for pleural effusion, and 412 images for pneumothorax. Table \ref{tab:dataset-distribution} presents the distribution of findings in the training and test datasets. There is no patient overlap between the training and test datasets.

\begin{table}[t]
\centering
\footnotesize
\caption{Distribution of positive findings in training and test sets of dataset $\mathcal{B}$. The training dataset consists of a total 16,953 images with at-least one abnormality. Images may have multiple findings, hence \% represent the proportion of total findings.}
\begin{tabularx}{\textwidth}{l *{3}{>{\centering\arraybackslash}X}}
\toprule
\textbf{Finding}  & \textbf{Positive Training Samples} & \textbf{Total Test images} & \textbf{Positive Test Samples} \\
\midrule
Consolidation    & 7266 (4.01\%)    & 333  & 141 (42.34\%)   \\
Atelectasis      &  128 (0.07\%)  & 333  & 83 (24.93\%)   \\
Pleural Effusion  & 5181 (2.86\%)    & 331  & 94 (28.34\%)   \\
Pneumothorax     & 4212 (2.32\%)    & 412  & 170 (51.21\%)   \\
\bottomrule
\end{tabularx}
\label{tab:dataset-distribution}
\end{table}

\subsubsection*{Dataset $\mathcal{C}$}
is the public VinDR-CXR dataset \cite{nguyen2022vindr}, comprising 18,000 CXR images manually annotated by a team of 17 experienced radiologists. Annotations include 22 local labels, provided as bounding boxes around abnormalities, and 6 global labels indicating suspected diseases. The dataset is split into 15,000 training, and 3,000 test images.

\begin{figure}[t]
    \centering
    \includegraphics[width=0.8\linewidth]{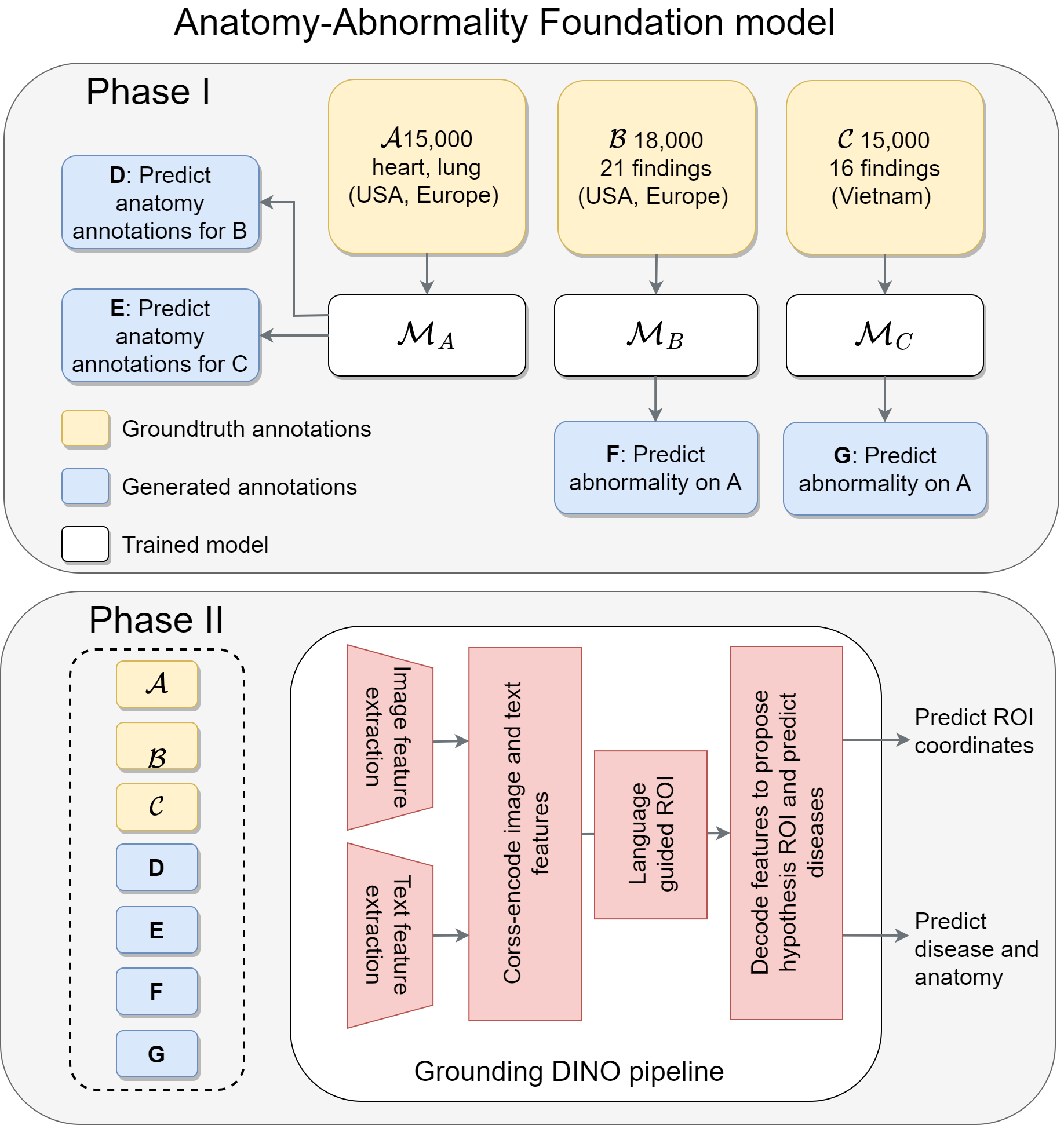}
    \caption{Overview of the student-teacher training process to create a foundation model with GDINO \cite{liu2024grounding}. A GDINO model is trained on each dataset with all available classes and used to generate weak annotations on remaining sets. A final model is trained in phase II with both gold-standard and generated annotations. The model outputs both predicted bounding boxes and classification scores for each box.}
    \label{fig:phase1_overview}
\end{figure}
\subsubsection*{Dataset $\mathcal{M}$}
is the public VinDR-Mammo dataset \cite{NguyenMammo}, comprising 5,000 full-field digital mammography (FFDM) exams, each with four standard views: craniocaudal (CC) and mediolateral oblique (MLO) for both left and right breasts. Each exam was independently double-read by two radiologists, with any discrepancies resolved by a third radiologist. The dataset includes breast-level annotations following the Breast Imaging Reporting and Data System (BI-RADS), covering assessment categories, breast density, and findings such as masses, calcifications, and architectural distortions. In particular, the dataset consists of images from a Vietnamese population with a typically high breast density (90\% of density C \& D), which renders the detection of mass a challenging task. There are a total of 15,000 images in the training dataset of which 50\% are normal images, and 4,000 images in the test dataset. The validation set is created as per previous studies \cite{ghosh2024mammo,emdetr} for a fair comparison. 

\section{Methods}\label{sec:method}

Grounding DINO (GDINO) \cite{liu2024grounding} serves as our baseline model and is illustrated in the second half of Fig.~\ref{fig:phase1_overview}. This model is an open-vocabulary object detector that aligns local image features with semantic textual or referring expressions. GDINO builds upon the GLIP \cite{li2022grounded} framework, which utilizes “grounding” phrase information to bootstrap the detection decoder training. Through this process, the model learns semantically rich representations that enable detection using text-based queries, achieved via contrastive learning between image regions and language embeddings. GDINO achieves impressive performance by pretraining the backbone and the detection encoder-decoder end-to-end on 1.8 million image-text bounding box annotations and captions, enabling the decoder to effectively associate a vast array of visual features with corresponding text embeddings. 

To approximate this level of performance within the medical domain, we initially explore the integration of anatomical and abnormality annotations from CXRs as a unified training set. Our objective was to explore whether increasing the number of medical annotations could improve the detection encoder-decoder's ability to identify and represent clinically important characteristics effectively. However, due to the sparsity of class-specific annotations and class imbalance, training an object detector by simply combining multiple non-overlapping datasets often results in catastrophic forgetting. This significantly degrades both classification and detection performance \cite{zhao2020object}. In the natural image domain, several studies \cite{zhao2020object,cai2022bigdetection} have demonstrated the effectiveness of pseudo-labelling techniques to unify label spaces across datasets with overlapping annotations. Conversely, the medical imaging domain, especially CXRs, present additional challenges. The anatomical and abnormality categories are inherently non-overlapping and share similar visual characteristics, leading to potential interference during training. This feature similarity limits the applicability of conventional pretraining and pseudo-labeling methods in the CXR domain. To address these challenges, we adopt a student–teacher training strategy, from Zhao et al. \cite{zhao2020object}, which effectively combines anatomical and abnormality datasets while mitigating catastrophic forgetting.

Fig.~\ref{fig:phase1_overview} depicts the training process of developing a foundation model with exclusive CXR images. First, individual models $\mathcal{M}_A,\mathcal{M}_B,\mathcal{M}_C$ are trained on their respective datasets $\mathcal{A},\mathcal{B},\mathcal{C}$ using domain-specific labels. These models then serve as teachers to generate pseudo-annotations for missing classes in the other datasets.  Specifically, for each dataset $\mathcal{D} \in \{\mathcal{A},\mathcal{B},\mathcal{C}\}$, the corresponding teacher models $\mathcal{M}_D$ with $\mathcal{D}' \neq \mathcal{D}$ predict annotations for classes absent in $\mathcal{D}$. Predictions with confidence scores above a class-specific threshold $\tau_c$ are retained as weak annotations, where $\tau_c$ is set to 95\% sensitivity on the Reciever-Operating-Curve (ROC) curve for class $c$ \cite{bhat2023aucreshaping}. For example, model $\mathcal{M}_C$ trained on dataset $\mathcal{C}$ predicts the 16 abnormalities on dataset $\mathcal{A}$, as well as non-overlapping abnormalities in dataset $\mathcal{B}$. This process repeats until all datasets contain annotations for all classes in $\mathcal{C}$, effectively unifying the label space. 

For anatomy predictions from $\mathcal{A}$ and $\mathcal{B}$, we select the top prediction per image and class without thresholding. We filter predictions by bounding box size and location to exclude those that are disproportionately large or small relative to the image, indicating significant Intersection-Over-Union (IoU) errors. Finally, a foundation model is trained using the combination of all annotations with both strong and weak annotations. We refer to this as the Anatomy-Abnormality Foundation Model (AAFM). 

\begin{figure}[t]
    \centering
    \includegraphics[width=0.8\linewidth]{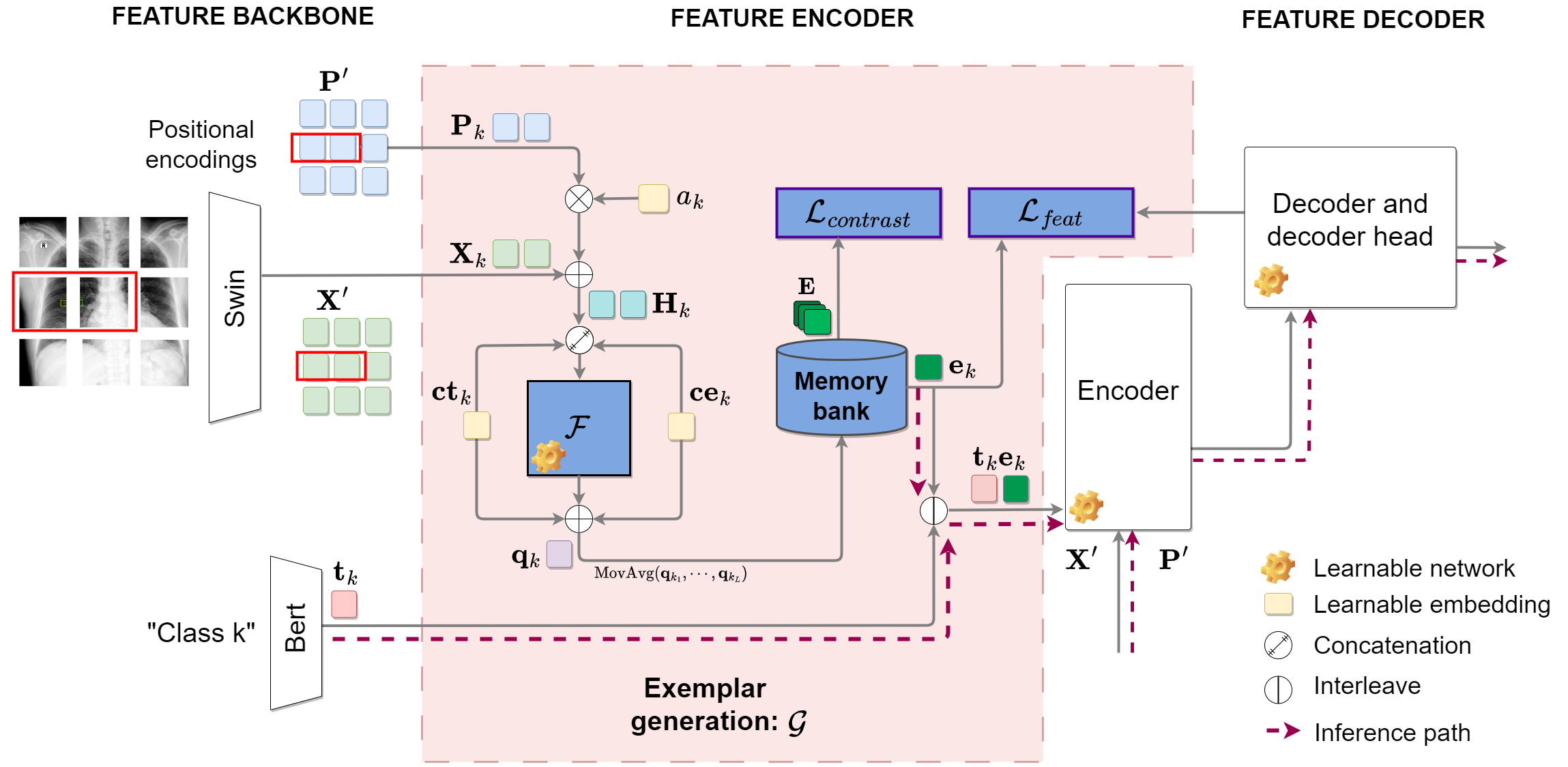}
    \caption{Overview of EM-DETR \cite{emdetr} detection. An exemplar feature embedding is generated for each class and interleaved with the text embedding. The detection encoder/decoder perform cross-attention between the image and interleaved embeddings, thereby training the decoder to search for classes based on their correlated text phrases as well as class-specific features.}
    \label{fig:emdetr_overview}
\end{figure}

Next, we incorporate annotations from other domains, which may share similar textual labels but differ visually, and use our previously introduced example-based SOTA method, EM-DETR \cite{emdetr}, to train a second foundation model that unifies the multi-domain representation space. Fig.~\ref{fig:emdetr_overview} shows the overview of EM-DETR \cite{emdetr}. EM-DETR is based on GDINO \cite{zhang2022dino,liu2024grounding}, guiding detection through ``class-specific exemplar embeddings'' along with language. These examples are based on visual characteristics obtained from the region of interest (annotated section) of certain categories, facilitating a ``example-based search''. The framework integrates exemplar generation, supplementary loss functions, and an ``iterative'' training approach to improve robustness and performance.

we use a frozen Swin \cite{liu2021swin} Transformer image backbone to extract shifted window patch embeddings $\mathbf{X}'$ and positional encodings $\mathbf{P}'$ for a given input image. For each class $k$, tokens $\mathbf{X}_k$ and $\mathbf{P}_k$ are selected using annotations, and are used to compute $\mathbf{H}_k$. This is mathematically denoted as:

\begin{equation}
\mathbf{H}_k = \mathbf{X}_k + a_k \times \mathbf{P}'
\quad \text{where } k \text{ denotes the class.}
\label{eq:hk}
\end{equation}
$a_k$ is a learnable scalar parameter that modulates the impact of the positional embeddings $\mathbf{P}_k$. The positional information provides crucial cues to the decoder, given that certain CXR findings exhibit global spatial patterns, while others are restricted to localized anatomical regions. $\mathbf{H}_k$ is passed to a trainable Transformer module $\mathcal{F}$, which generates the representative class-specific feature embedding prototype $\mathbf{q}_k$, expressed as: 
\begin{equation}
\mathbf{q}_k = \mathcal{F}\big(\mathrm{concat}(\mathbf{ct}_k, \mathbf{ce}_k, \mathbf{H}_k)\big) + \mathbf{ct}_k + \mathbf{ce}_k, 
\label{eq:qk}
\end{equation}
where $\mathbf{ct}_k$ and $\mathbf{ce}_k$, are class-specific learnable token and positional embeddings. To stabilize learning, a moving average of $\mathbf{q}_k$ is computed and stored in a ``memory bank'' as $\mathbf{e}_k$. Exemplar embeddings are interleaved with text embeddings to facilitate the creation of hypothesis regions by the decoder. These regions are formed by contrasting the correlation between image embeddings and the interleaved embeddings. Two additional losses called $\mathcal{L}_{contrast}$ and $\mathcal{L}_{feat}$ are also introduced. These losses enforce orthogonality between class exemplar embeddings $\mathbf{e}_k$  and ensure the decoder's top proposal aligns with the respective $\mathbf{e}_k$, consequently ensuring class-specific uniformity.

Although these mechanisms define how the model learns class-specific representations, its effectiveness also depends on the training strategy and dataset composition. The contrastive loss learning scheme affects the model's performance based on training class selection. Furthermore, the characteristics of various normal anatomical structures are similar to diverse findings. Thus, an iterative training approach is adopted, where the model is incrementally improved based on the output of each stage. The first stage foundation model employs all the annotations of dataset $\mathcal{B}$, $\mathcal{C}$ and $\mathcal{M}$. 

\begin{figure}[t]
    \centering
    \includegraphics[width=0.8\linewidth]{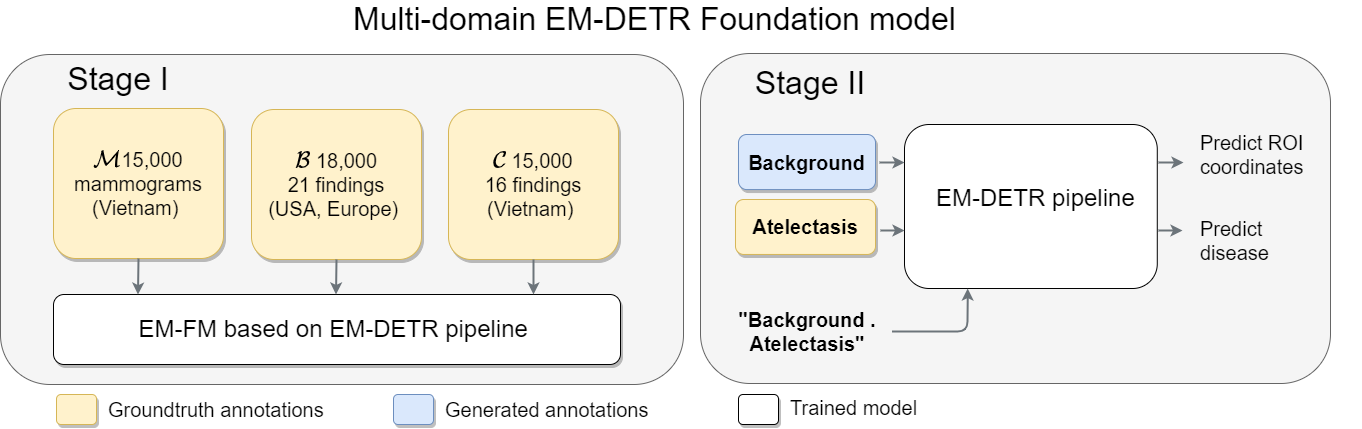}
    \caption{Overview of training the EM-DETR based foundation model, called EM-FM. In the first stage, all available radiology datasets are used to pretrain the detection encoder/decoder pair. The decoder is trained to detect medical features that correlate to specific text embeddings. In the second stage, the weights are refined for class-specific detection using a small subset of foreground vs. background annotations. The background annotations are generated based on data availability \cite{emdetr}.}
    \label{fig:em_fm_overview}
\end{figure}

In the second stage, model weights are refined using a per-class background-versus-foreground dataset. The background annotations are generated by selecting regions from the known normal images. To select suitable background regions, we utilize a set of known locations of the findings within the training dataset. This process aims to capture the normal anatomies that must be distinguished for those specific findings. During inference the annotated regions are not available, and the stored $\mathbf{e}_k$ are used to guide detection. This is indicated as dotted lines in Fig.~\ref{fig:emdetr_overview}. This model is referred to as the EM-DETR Foundation Model (EM-FM). $\mathcal{A}$ was excluded in EM-FM, as combining multiple domains resulted in a large number of annotations, reducing the need for auxiliary anatomical supervision. 

\subsection{Evaluation criteria}

We evaluate image-level classification and localization performance for baseline, SOTA, and foundational models. For classification, the maximum predicted probability for each class within an image is recorded to compute the area under the ROC curve (AUC) score. Localization accuracy is assessed using mean Average Precision at 50\% IoU $mAP_{50}$, following the COCO protocol in MMDetection \cite{chen2019mmdetection}. We also assess the model’s Free-response Receiver Operating Characteristic (FROC) curve by measuring sensitivity across false positive (FP) rates. The FROC curve represents a more practical metric for diagnostic lesion detection systems, where the precise delineation of the boundaries of complex findings \cite{pham2022accurate,9395510,ghesu2022contrastive,7837683} is contentious. Sensitivity is calculated based on the distance between the centers of predicted regions and groundtruth annotations.
\begin{table}[t]
\footnotesize
\centering
\caption{Comparison of the image-level AUC classification scores, where $*$ shows statistical significance of $p<0.05$ when AAFM and EM-FM are compared to the Reference Product Solution (RPS) \cite{ghesu2022contrastive}. $\varepsilon$ indicates a ResNet-101 backbone, while $\dagger$ indicates a Swin-L architecture. All results have a standard deviation $\epsilon < \pm0.005$.}
\begin{tabularx}{\textwidth}{l@{\hskip 8pt}Y@{\hskip 8pt}Y@{\hskip 8pt}Y@{\hskip 8pt}Y}
\toprule
Finding 
& \multicolumn{1}{c}{RPS\textsuperscript{$\varepsilon$}} 
& \multicolumn{1}{c}{Baseline\textsuperscript{$\dagger$}} 
& \multicolumn{1}{c}{AAFM\textsuperscript{$\dagger$}} 
& \multicolumn{1}{c}{EM-FM\textsuperscript{$\dagger$}}\\
\midrule
Consolidation      & 0.932 & 0.926 & \textbf{0.943}* & 0.921 \\
Atelectasis        & 0.949 & 0.959 & \textbf{0.974}* & 0.932 \\
Pleural effusion   & 0.993 & \textbf{0.994} & 0.993 & 0.993\\
Pneumothorax       & 0.975 & \textbf{0.978} & \textbf{0.978} & 0.968 \\
\midrule
Average            & 0.962 & 0.959 & \textbf{0.972}* & 0.954 \\
\bottomrule
\end{tabularx}

\label{tab:auc}
\end{table}

\section{Experimental Results}
\label{sec:resutl}

\subsection{Experimental Setup}
Experiments were conducted on a single node with four 40 GB A100 GPUs, within the MMDetection framework \cite{chen2019mmdetection}. Default data augmentations and hyper-parameters were applied. The Swin-L backbone used a per-GPU batch size of 1, with a learning rate of 0.0008. Standard CXR-related augmentations such as random auto-contrast, sharpness adjustments, and random horizontal flipping, were added to the default augmentation pipeline. Furthermore, additional terms such as ``right lung'' and ``left lung'' were added to the text vocabulary, allowing the creation of distinct tokens for the intended left and right categories. Predictions from the anatomy teacher below 0.8 score and deviations greater than 50\% from the average size of the ground truth were excluded. For mammogram images, the background dark regions were cropped, and the annotations were recalculated for the resized images. Each experiment was conducted five times, training upto 25 epochs. The reported results correspond to the mean performance calculated from the top 5 check points selected based on $mAP_{50}$. 

\subsection{Comparison with SOTA}

Table~\ref{tab:auc} presents the image-level AUC results for four key CXR findings. It compares our method against the commercial Reference Product Solution (RPS) \cite{ghesu2022contrastive}, which is based on ResNet-101 FCOS \cite{tian2020fcos} solution, as well as the baseline GDINO model. The RPS and baseline is trained solely on Dataset $\mathcal{B}$, AAFM and EM-FM are trained on $\mathcal{A}$-\textbf{G} and $\mathcal{B},\mathcal{C},\mathcal{M}$ respectively. We observe that AAFM outperforms SOTA with a statistically significant 1\% point improvement in the average AUC ($p < 0.05$) score. This gain can be attributed to the effective integration of multiple datasets and the increased model capacity. Notably, when datasets are naively combined, the classification score for pneumothorax drops to 0.75, highlighting the importance of our integration strategy. The RPS \cite{ghesu2022contrastive}, while demonstrating strong performance, benefits from large-scale pretraining on 100 million images and relies on Non-Maximum Suppression (NMS) for post-processing. 

\begin{table}[t]
\footnotesize
    \centering
     \caption{Comparison of the average detection $mAP_{50}$ results between the baseline, AAFM, EM-DETR and EM-FM. $\dagger$ indicates a Swin-L image backbone and $*$ shows statistical significance with $p<0.05$. All results have a standard deviation $\epsilon < \pm0.1$.}
   \begin{tabularx}{\textwidth}{l@{\hskip 8pt}Y@{\hskip 8pt}Y@{\hskip 8pt}Y@{\hskip 8pt}Y}
\toprule
Findings in CXRs
& \multicolumn{1}{c}{Baseline\textsuperscript{$\dagger$}} 
& \multicolumn{1}{c}{AAFM\textsuperscript{$\dagger$}}
& \multicolumn{1}{c}{EM-DETR\textsuperscript{$\dagger$}}
& \multicolumn{1}{c}{EM-FM\textsuperscript{$\dagger$}} \\
\midrule
Consolidation     & 0.450 & 0.386* & 0.539* & \textbf{0.596}* \\
Atelectasis        & 0.363 & 0.\textbf{366} & 0.346 & 0.363 \\
Pleural effusion    & 0.244 & 0.240 & 0.715* & \textbf{0.756}* \\
Pneumothorax       & 0.419 & 0.444* & 0.508* & \textbf{0.556}* \\
\midrule
Average & {0.369} & {0.359} & 0.527 & \textbf{0.567}* \\
\midrule
Findings in Mammography & & & \\
\midrule
Mass  & 0.583 & --- & 0.699* & \textbf{0.721}* \\ 
       \bottomrule
    \end{tabularx}
   
    \label{tab:map}
\end{table}

Table.~\ref{tab:map} compares the $mAP_{50}$ of the foundation models with the baseline and previous SOTA EM-DETR across CXR and mammogram findings. Compared to the baseline, EM-FM achieved substantial and statistically significant gains in CXR detection, with improvements of 14.6, 51.2 and 13.7\% points for consolidation, pleural effusion and pneumothorax, respectively. This translates to an average of 19.9\% point increase in $mAP_{50}$ (0.567 vs.\ 0.369, $p<0.05$). Compared to EM-DETR, EM-FM improved $mAP_{50}$ by 2-6\% points across these findings, and also outperformed EM-DETR in mammography by 2.2\% points. 

The limited gains in atelectasis detection are attributed to inherent labelling ambiguity stemming from clinical overlap with conditions like consolidation and distalectasis. These findings often share similar radiological features and may represent different stages along a pathological spectrum, leading to noisy or inconsistent groundtruth annotations across datasets. On the other hand, although AAFM underperforms on atelectasis and pleural effusion, relative to the baseline, these differences are not statistically significant. This gap stems from classification-optimized confidence thresholds in student-teacher training and variability in GDINO’s contrastive learning, which depends on the chosen classes for evaluation. 

\begin{figure}[t]
    \centering
    \includegraphics[width=0.8\linewidth]{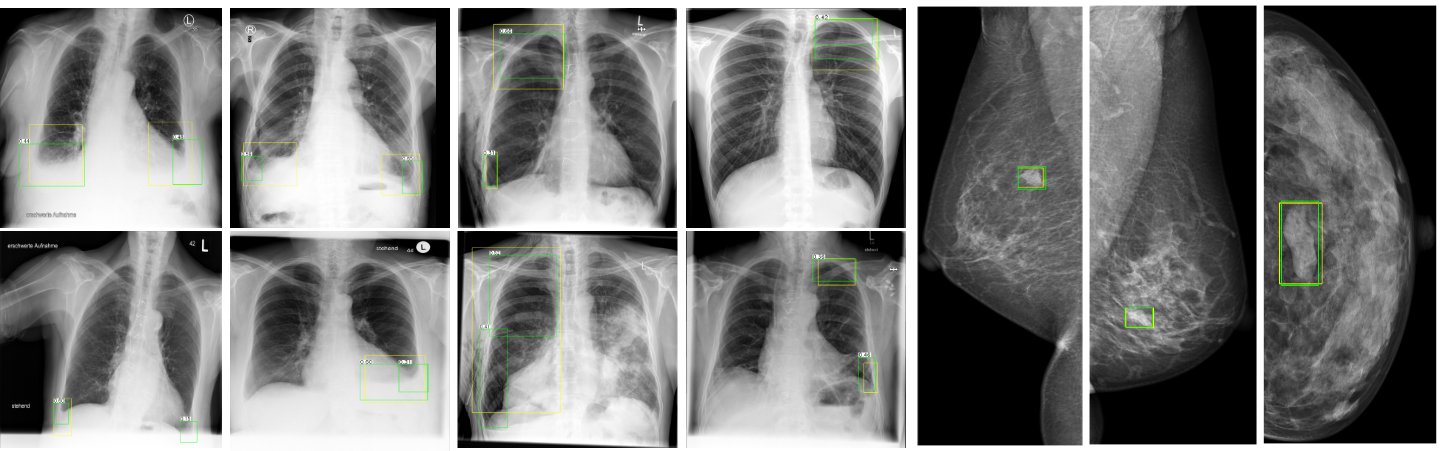}
    \caption{Qualitative examples of bounding box predictions for pleural effusion (first four) and pneumothorax (next four) in CXR images, followed by mass predictions in mammograms. Groundtruth in yellow and predictions in green with their prediction scores are also shown.}
    \label{fig:qual_examples}
\end{figure}

Table. \ref{tab:froc} compares the sensitivity at 0.5 FP between the RPS \cite{ghesu2022contrastive}, baseline, and foundation models for four CXR findings. While the RPS leads with an average sensitivity of 0.954, EM-FM remains competitive at 0.927, despite a significant drop in atelectasis sensitivity. AAFM has significant lower sensitivity for atelectasis but excels in pneumothorax, while the naive combination of datasets yields a sensitivity of 0.18 for pneumothorax. These results demonstrate effective multi-dataset integration with EM-FM maintaining robust detection performance. Fig.~\ref{fig:qual_examples} showcases qualitative samples of pleural effusion, pneumothorax on CXR images, and mass predictions in mammograms. We observe that the predictions produce smaller, more precise bounding boxes that align closely with the annotated regions.

\begin{table}[t]
\footnotesize
\centering
\caption{Comparison of average sensitivity at 0.5 FP between RPS \cite{ghesu2022contrastive}, baseline, AAFM and EM-FM. $\varepsilon$ indicates a ResNet-101 backbone, while $\dagger$ indicates a Swin-L architecture. All results have a standard deviation $\epsilon < \pm0.01$.}
\begin{tabularx}{\textwidth}{l@{\hskip 8pt}Y@{\hskip 8pt}Y@{\hskip 8pt}Y@{\hskip 8pt}Y}
\toprule
Finding 
& \multicolumn{1}{c}{RPS\textsuperscript{$\varepsilon$}}
& \multicolumn{1}{c}{Baseline\textsuperscript{$\dagger$}} 
& \multicolumn{1}{c}{AAFM\textsuperscript{$\dagger$}} 
& \multicolumn{1}{c}{EM-FM\textsuperscript{$\dagger$}} \\
\midrule
Consolidation      & \textbf{0.900} & 0.840 & 0.740 & 0.891 \\
Atelectasis        & 0.950 & \textbf{0.970} & 0.940 & 0.896\\
Pleural Effusion   & \textbf{0.997} & 0.987 & 0.987 & 0.987 \\
Pneumothorax       & 0.970 & 0.945 & \textbf{0.973} & 0.935 \\
\midrule 
Average & \textbf{0.954} & 0.936 & 0.910 & 0.927 \\
\bottomrule
\end{tabularx}
\label{tab:froc}
\end{table}

\begin{figure}
    \centering
    \includegraphics[width=0.75\linewidth]{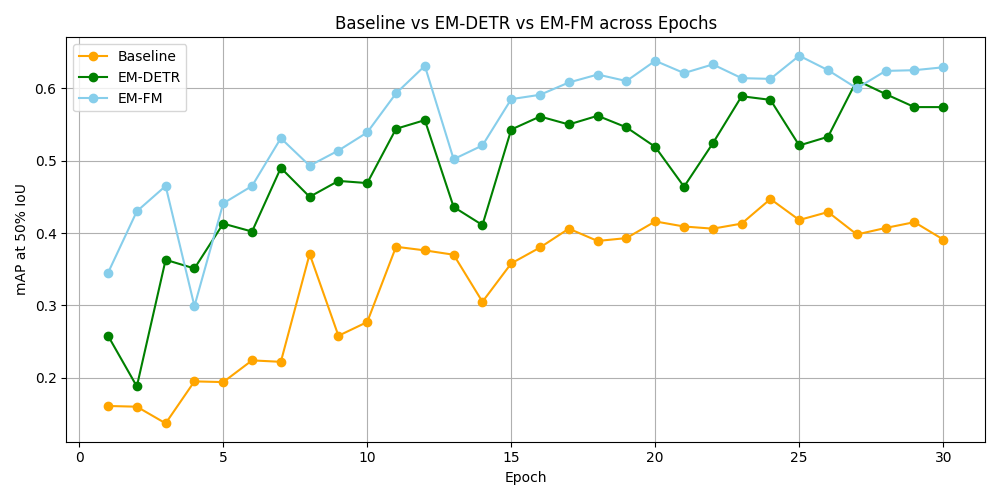}
    \caption{Comparison of per-epoch pneumothorax $mAP_{50}$ values for the baseline and EM-FM methods. The EM-FM model demonstrates faster and more stable convergence compared to the baseline and EM-DETR.}
    \label{fig:convergence}
\end{figure}

As shown in Fig. \ref{fig:convergence} , leveraging foundation models not only improves performance but also accelerates and stabilizes training convergence compared to the baseline. The EM-FM model reaches higher $mAP$ values overall, demonstrating the benefits of pretrained visual representations for faster optimization. This observation aligns with prior large-scale pretraining literature. It is important to note that the baseline performance can exhibit considerable variability. This instability primarily arises from its reliance on contrastive learning, where the model attempts to simultaneously optimize for both detection and contrastive objectives across multiple classes. Consequently, the performance is highly dependent on the specific set of classes used during training, as the model must balance the representation of each class while learning discriminative features. This intrinsic sensitivity to class selection can lead to fluctuations in baseline results, particularly when the dataset contains classes with differing levels of difficulty or prevalence. Conversely, owing to their example-based learning paradigm, the EM-DETR and EM-FM models generally exhibit more stable performance when trained on a large number of classes. This stability arises from their ability to leverage learned feature representations to match examples, rather than relying solely on class-dependent contrastive language objectives. As a result, these models are less sensitive to class selection and distribution, enabling more consistent detection performance across diverse and heterogeneous datasets.

\section{Conclusion}
\label{sec:conclusion} 

This study explores creating a detection foundation model for all radiography images across multiple modalities. We first demonstrate that the combination of disjoint and overlapping set of anatomy and abnormality datasets, achieves SOTA classification performance. In contrast to their naive combination, we prevent catastrophic forgetting by weakly labelling missing annotations using a student-teacher approach. While not specifically optimizing for localization, we achieve comparable capabilities for most conditions, with the exception of atelectasis. We then incorporate all available CXR and mammography data into our previously proposed example-based detection model EM-DETR, resulting in significant detection improvements over prior SOTA across both modalities. Leveraging foundation models also facilitates faster and more stable training convergence, as widely reported in foundation model research. Future work will extend this approach to additional findings and datasets to further validate its generalizability and robustness.


\begin{credits}
\subsubsection{\ackname} 
This study was funded by Siemens-Healthineers.

\subsubsection{\discintname}
The authors have no competing interests to declare.

\end{credits}
%
%
%
\bibliographystyle{splncs04}
\bibliography{cvip}

\end{document}